\documentclass[runningheads]{llncs}
\usepackage[T1]{fontenc}
\usepackage{graphicx}
\usepackage{color}
\usepackage{hyperref}

\usepackage{amsmath}
\usepackage{amssymb}
\usepackage{mathtools}

\usepackage[capitalize,noabbrev]{cleveref}

\usepackage{amssymb}
\usepackage{amsfonts}
\usepackage{amsmath}
\usepackage{amsxtra}
\usepackage{cancel}
\usepackage{dsfont}
\usepackage{graphicx}
\usepackage{tikz}
\usepackage{tikz-qtree}
\usetikzlibrary{shapes}
\usetikzlibrary{positioning}
\usetikzlibrary{trees}
\usepackage{mathcomp}
\usepackage{mathtools}
\usepackage{multirow}
\usepackage{verbatim}
\usepackage{polynom}
\usepackage{textcomp}
\usepackage{float}
\usepackage{pdflscape}
\usepackage{csquotes}
\usepackage{afterpage}
\usepackage{makecell}
\usepackage{listings}
\usepackage{url}
\usepackage{enumitem}
\usepackage{minibox}
\usepackage{algorithm}
\usepackage{algorithmic}
\usepackage{shuffle}
\usepackage{svg}
\usepackage{pifont}
\usepackage{subcaption}
\usepackage{xspace}
\usepackage{siunitx}
\usepackage{booktabs}

\DeclareMathSymbol{\mlq}{\mathord}{operators}{``}
\DeclareMathSymbol{\mrq}{\mathord}{operators}{`'}

\newcommand{\R}{\mathbb R}
\newcommand{\N}{\mathbb N}

\newcommand{\1}{\mathds{1}}

\renewcommand{\phi}{\varphi}
\renewcommand{\Im}{\operatorname{Im}}

\newcommand{\norm}[1]{\left\lVert#1\right\rVert}
\newcommand{\abs}[1]{\left| #1 \right|}

\newcommand{\softmax}{\operatorname{softmax}}

\newcommand{\tensor}{\otimes}

\newcommand{\bx}{\mathbf{x}}

\renewcommand{\O}{\mathcal{O}}

\newcommand{\taylorsm}{\operatorname{T-SM}}
\newcommand{\ops}{\operatorname{ops}}
\newcommand{\entr}{\operatorname{entries}}

\newcommand{\name}{TaylorShift\xspace}

\begin{document}
\title{\name: Shifting the Complexity of Self-Attention from Squared to Linear (and Back) using Taylor-Softmax}
\titlerunning{\name} %
\author{Tobias Christian Nauen\inst{1,2} \and
Sebastian Palacio\inst{2} \and
Andreas Dengel\inst{1,2}}
\authorrunning{T. Nauen et al.}
\institute{
	Department of Computer Science, RPTU Kaiserslautern-Landau, Gottlieb-Daimler-Straße, Kaiserslautern, Germany \and
	German Research Center for Artificial Intelligence (DFKI), Trippstadter Str. 122, Kaiserslautern, Germany \\ \email{\{tobias\_christian.nauen,sebastian.palacio,andreas.dengel\}@dfki.de}
}
\maketitle              %
\begin{abstract}
The quadratic complexity of the attention mechanism represents one of the biggest hurdles for processing long sequences using Transformers.
Current methods, relying on sparse representations or stateful recurrence, sacrifice token-to-token interactions, which ultimately leads to compromises in performance.
This paper introduces \name, a novel reformulation of the Taylor softmax that enables computing full token-to-token interactions in linear time and space.
We analytically determine the crossover points where employing \name becomes more efficient than traditional attention, aligning closely with empirical measurements.
Specifically, our findings demonstrate that \name enhances memory efficiency for sequences as short as 800 tokens and accelerates inference for inputs of approximately 1700 tokens and beyond.
For shorter sequences, \name scales comparably with the vanilla attention.
Furthermore, a classification benchmark across five tasks involving long sequences reveals no degradation in accuracy when employing Transformers equipped with \name.
For reproducibility, we provide access to our code under \url{https://github.com/tobna/TaylorShift}.

\end{abstract}

\section{Introduction}
Ever since their introduction by Vaswani et al. \cite{Vaswani2017}, Transformers have revolutionized numerous domains of deep learning, from Natural Language Processing to Computer Vision, while also underpinning the emergence of novel applications such as Large Language Models.
This success stems largely from their ability to capture intricate dependencies and token-to-token interactions.

To extend the utility of Transformers to more complex tasks, they need to be able to process long sequences.
However, the computational complexity of the attention mechanism scales quadratically in the length of the input sequence $\O(N^2)$. Therefore, computing twice as many sequence elements requires four times the number of computations, which hinders scaling to very long context windows.
This makes some practitioners turn to approaches like compressing portions of the input into single states \cite{Bulatov2023,Dai2019}, which reduces the amount of information available at each step.
Despite this progress, exploiting long context windows to significantly improve performance and incorporate new information without retraining remains challenging.
Current Transformers encounter limitations when processing long documents, high-resolution images, or a combination of data from multiple domains and modalities.
Especially, considering the limited resources of smaller enterprises or individual consumers.

While linearly scaling Transformers have been proposed, these often experience compromised accuracy \cite{Nauen2023}, specialize in a particular domain, like language \cite{Zaheer2020} or images \cite{Liu2021}, or only convey averaged global information across tokens, neglecting individual token-to-token interactions \cite{Babiloni2023,ElNouby2021}.
These models end up being ill-suited for handling longer sequences, leaving the standard Transformer as the preferred choice due to its large capacity and established performance \cite{Lin2022}. 

In this work, we approach this bottleneck of the Transformer by reformulating the softmax function in the attention mechanism after introducing the Taylor approximation of the exponential.
While some methods alter the softmax, their goal is to split interactions of queries and keys, computing global average interactions only \cite{Babiloni2023,Choromanski2021}.
In contrast, our proposed approach, \name, preserves individual token-to-token interactions.
Combining a tensor-product-based operator with the Taylor approximation of the exponential function allows us to compute full token-to-token interactions in linear time.
Moreover, this approach has the added benefit of adhering to concrete error bounds when viewed as an approximation of vanilla attention \cite{Keles2023}.
We show that a naive implementation of this linearization is numerically unstable and propose a novel normalization scheme that enables its practical implementation.
For short sequences, \name can default back to quadratic scaling to preserve efficiency.
We apply \name to a diverse set of tasks on images, text, and mathematical operations.

Our paper starts with the related work (\Cref{sec:related_work}), providing context for our contributions. 
In \Cref{sec:taylor_softmax}, we introduce two implementations of \name, efficient for short and long sequences, respectively, and our novel normalization scheme. 
Beyond the $\O$-notation, we delve into the efficiency analysis of \name, identifying specific conditions where it excels, both theoretically (\Cref{sec:theoretical_analysis}) and empirically (\Cref{sec:experiments}).
Finally, we conclude in \Cref{sec:conclusion}.

\section{Related Work}
\label{sec:related_work}

To contextualize \name, we review work on efficient attention, how Taylor approximations are used in ML, and their application for attention specifically.

\paragraph{Linear Complexity Attention}
Various strategies have been proposed to devise attention mechanisms with linear complexity.
Sparse attention mechanisms, like Swin \cite{Liu2021} (images) or BigBird \cite{Zaheer2020} (text), only selectively enable token-to-token interactions and their effectiveness heavily depends on the input modality.
Kernel-attention methods \cite{Choromanski2021,Babiloni2023}, decouple the influence of queries and keys, leading to a global average transformation instead of individual token-to-token interactions.
Mechanisms like Linformer \cite{Wang2020} apply transformations on the sequence direction, restricting them to a specific input size.
For a comprehensive exploration of this topic, readers are referred to \cite{Fournier2023,Nauen2023}.
While these linear attention mechanisms offer innovative solutions to computational challenges, their performance nuances and lack of adaptability compared to \name warrant further exploration.

\paragraph{Taylor Approximation in ML}
Applying Taylor approximations has proven to be a powerful technique in deep learning.
In Explainable AI, the Deep Taylor Decomposition \cite{Montavon2015} employs a linear Taylor decomposition of individual neurons to propagate the relevancy of each part of the input.
Linear Taylor approximations also are utilized in network pruning, where they are leveraged to quantify the influence of individual neurons on a loss value \cite{Gaikwad2018,Molchanov2017}.
TE-CSR \cite{Xing2020} directly utilizes a Taylor series to gather multivariate features in the domain of image fusion.
Recently, TaylorNet \cite{Zhao2023a} and Taylorformer \cite{Nivron2023} treat the factors of a Taylor series, as learnable parameters.
The Taylor softmax \cite{Vincent2015}, introduced to enable efficient calculation of loss values, outperformed the traditional softmax in image classification \cite{Brebisson2016}.
In this work, we leverage insights from these diverse applications of Taylor series to enable the efficient calculation of attention.

\paragraph{Taylor Approximation in Attention}
Recently, \cite{Qiu2023} 
adopt the first order Taylor softmax in the attention mechanism.
However, this is limited to linear token-interactions.
To emphasize local interactions, they add a convolution operation.
In contrast, we compute individual non-linear interactions in linear time.

\cite{Keles2023}, an analysis of efficient attention mechanisms, mentions the theoretical possibility of leveraging higher order Taylor softmax to approximate the attention mechanism in linear time, but with exponential complexity in the order of the Taylor approximation.
In this work, we draw inspiration from this theoretical analysis and develop a viable, working implementation based on Taylor series.
We analyze the efficiency gains beyond the $\O$-notation, estimating transition points where it outperforms standard attention.

\section{\name}
\label{sec:taylor_softmax}

This section describes the formal derivation of \name and its algorithmic implementation.
Starting from a direct, non-efficient formulation, we proceed to mathematically derive a provably efficient alternative.
An investigation into scaling behaviors will lead to the incorporation of a novel normalization scheme.

\subsection{Direct \name}
Taylor-Softmax approximates the softmax's exponential function by its $k$-th order Taylor approximation:
\begin{align*}
	\exp(x) \approx \sum_{n = 0}^k \frac{x^{n}}{n!}.
\end{align*}

For a vector $\bx \in \R^d$, with Hadamard powers $\bx^{\odot n}$:
\begin{align*}
	\softmax(\bx) = \text{normalize}(\exp \bx) \approx \text{normalize}\left( \sum_{n = 0}^k \frac{\bx^{\odot n}}{n!} \right) =: \taylorsm^{(k)}(\bx) 
\end{align*}
Here, the $\text{normalize}$ operation is division by the $\ell^1$-norm: ${\bx \mapsto \frac{\bx}{\sum_i \abs{\bx_i}}}$.
For even $k$, Taylor-Softmax generates a probability distribution, since it is positive
and its terms sum to one.
$k = 2$ balances computational cost and expressivity \cite{Brebisson2016}.

\label{sec:trivial_implementation}
By using Taylor-Softmax, the attention mechanism for the query, key, and value matrices $Q, K, V \in \R^{N \times d}$, where $N$ is the length of the sequence and $d$ is the internal dimension, takes the form
\begin{align}\label{eq:taylor-attention}
	Y = \taylorsm \left(Q K^\top\right) V
\end{align}
with row-wise Taylor-Softmax.
We refer to the direct implementation of \Cref{eq:taylor-attention}, which calculates the large $N \times N$ attention matrix 
$\taylorsm \left(QK^\top \right)$ 
of token-to-token interactions before multiplying it by $V$, as \emph{direct-\name}.

\subsection{Efficient \name}
\label{sec:efficient_implementation}
Since direct-\name does not scale well, we derive a more efficient implementation.
We can archieve this, by splitting the influence of the taylor approximation of the exponential function among the matrices $Q$ and $K$ and pushing the normalization operation to the end, after multiplying by $V$.
Mathematically the result will still be the same, but by switching up the order of operations, the computational complexity can be reduced from $\O(N^2 d)$ to $\O(Nd^3)$.

First, we rewrite the normalization operation by splitting it into nominator and denominator:
\begin{align*}
	Y_\text{nom} &= [1 + Q K^\top + \frac 1 2 (Q K^\top)^{\odot 2}] V, \\
	Y_\text{denom} &= [1 + Q K^\top + \frac 1 2 (Q K^\top)^{\odot 2}] \1_N, \\
	\Rightarrow Y &= Y_\text{nom} \oslash Y_\text{denom},
\end{align*}
where $\1_N \in \R^N$ is the vector of ones and ${}^{\odot 2}$ and $\oslash$ are the Hadamard power and division.
This representation allows us to disentangle the influence of the linear, squared, and constant terms of the Taylor approximation into their influence on $Q$ and $K$, respectively.

The constant and linear influence $[ 1 + QK^\top ] V = Q (K^\top V) + \Sigma_\text{col} V$ can trivially be computed in $\O(Nd^2)$, leaving us with $(Q K^\top)^{\odot 2} V$.
To handle this term efficiently, we define a tensor product on the internal dimension $d$:
\begin{align*}
	\boxtimes&: \R^{N \times d} \times \R^{N \times d} \to \R^{N \times d^2} \\
	[A \boxtimes B]_n &= \iota(A_n \tensor B_n) \in \R^{d^2} \hspace{10pt} \forall n = 1, ..., N
\end{align*}
Here, $A_n, B_n \in \R^d$, and $[A \boxtimes B]_n$ are the $n$-th entries of $A, B,$ and $A \boxtimes B$ respectively, $\otimes$ is the outer product of vectors\footnote{We identify 
the basis ${\{ e_i \otimes e_j \}_{ij}}$ of tensor space with the canonical basis $\{ e_{ij} \} \subset \R^{d \times d}$ of matrix space, viewed as a vector space. $\{ e_i \}_i$ is the canonical basis of $\R^d$.}, and ${\iota: \R^{d \times d} \xrightarrow{\sim} \R^{d^2}}$ is the canonical isomorphism of reordering the entries of a matrix into a vector.
This reordering operation can be described by a bijective map $\pi: \{1, ..., d\} \times \{1, ..., d\} \to \{1, ..., d^2\}$.
We define ${A^{\boxtimes 2} := A \boxtimes A}$.
Then we have $[A^{\boxtimes 2}]_{n, \pi(k, \ell)} = A_{n, k} A_{n, \ell}$.
This lets us linearize $(Q K^\top)^{\odot 2}$ by using the tensor operator $\boxtimes$ to unroll the square of a $d$-element sum along a sum of $d^2$ elements.
At position $ij$, we have
\begingroup
\allowdisplaybreaks
\begin{align*}
	[(Q K^\top)^{\odot 2}]_{ij} &= \left(\sum_{k = 1}^{d} Q_{ik} K_{jk}\right)^2 = \sum_{k, \ell = 1}^{d} Q_{ik} Q_{i\ell} K_{jk} K_{j\ell} \\
	&= \sum_{k, \ell = 1}^{d} [Q_{i} \otimes Q_{i}]_{k, \ell} [K_{j} \tensor K_{j}]_{k, \ell} = \sum_{k, \ell = 1}^{d} [Q^{\boxtimes 2}]_{i, \pi(k, \ell)} [K^{\boxtimes 2}]_{j, \pi(k, \ell)} \\
	&= [Q^{\boxtimes 2}]_i [K^{\boxtimes 2}]_j^\top. \\
\end{align*}
\endgroup
for $i,j = 1, ..., N$. And therefore
\begin{align} \label{eq:a_mod_def}
	\Rightarrow Y_\text{squ} := (Q K^\top)^{\odot 2} V = \underbrace{Q^{\boxtimes 2}}_{N \times d^2} \underbrace{(K^{\boxtimes 2})^\top V}_{=:A_\text{mod}}
\end{align}
This can be calculated in linear time in $N$ by multiplying from right to left.
Adding both the linear and the constant terms to the square-term gives:
\begin{align} \label{eq:efficient-implementation}
	Y_\text{nom} = \frac 1 2 Q^{\boxtimes 2} \left( (K^{\boxtimes 2})^\top V \right) + Q (K^\top V) + \Sigma_\text{col} V.
\end{align}
We calculate the nominator $Y_\text{nom}$ and denominator $Y_\text{denom}$ simultaneously using \Cref{eq:efficient-implementation} by setting 
$V \gets \left(\1_N \circ V \right) \in \R^{N \times (d+1)}$,
where $\circ$ is the concatenation operation.
The result ${\hat Y \in \R^{N \times (d+1)}}$ can then be split back into $Y_\text{denom} \in \R^N$ and $Y_\text{nom} \in \R^{N \times d}$ to get the final output:
\begin{align} \label{eq:eff_taylor_out}
	Y = \left[ \frac{[Y_\text{nom}]_{1 :}}{[Y_\text{denom}]_1}, ..., \frac{[Y_\text{nom}]_{N :}}{[Y_\text{denom}]_N} \right] \in \R^{N \times d}.
\end{align}

\begin{figure}[t]
	\centering
	\resizebox{.75\textwidth}{!}{
		\includegraphics{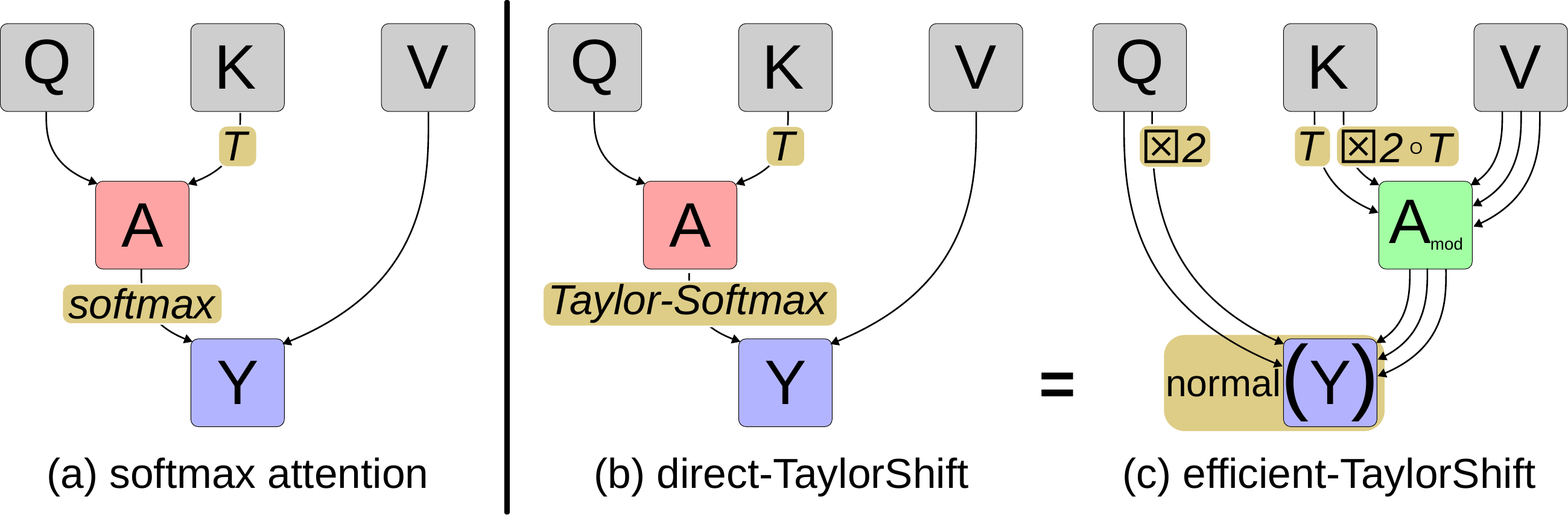}}
	\caption{Order of operations in softmax attention, direct-, and efficient-\name. Multi-paths for efficient-\name show squared, linear, and constant influence.}
	\label{fig:taylor_shift_operations}
\end{figure}

We refer to the result of \Cref{eq:eff_taylor_out} when calculating $\hat Y = Y_\text{denom} \circ Y_\text{nom}$ using \Cref{eq:efficient-implementation} as \emph{efficient-\name}.
\Cref{fig:taylor_shift_operations} visualizes the differences between direct- and efficient-\name.
The output of direct- and efficient-\name is the same mathematically, but the later scales linearly in $N$.

\begin{table}[t]
	\centering
	\caption{Mean size of intermediate expressions in efficient-\name, when rows of $Q, K,$ and $V$ are sampled uniformly from the unit sphere.}
	\label{tbl:scaling_behavior}
	\begin{tabular}{l@{\hskip 8pt}c@{\hskip 8pt}c@{\hskip 8pt}c@{\hskip 8pt}c@{\hskip 8pt}c}
		\toprule
		Expr. & $(K^{\boxtimes 2})^\top V = A_\text{mod}$ & $(QK^T)^2V$ & $QK^\top V$ & $Y_{denom}$ & $Y$ \\
		\cmidrule(r){1-1} \cmidrule(l){2-6}
		Size & $\frac{N+1}{\sqrt{d}}$ & $\frac{N}{d}$ & $\sqrt{N} \frac{4d + 1}{4 d}$ & $N \frac{d + 2}{2d}$ & $\sqrt{\frac{d}{N}}$ \\
		\bottomrule
	\end{tabular}
\end{table}

\begin{algorithm}[t]
	\caption{Efficient-\name with normalization}
	\label{alg:taylored_attention}
	\begin{algorithmic}[1]
		\REQUIRE Queries, Keys and Values $Q, K, V \in \R^{N \times d}$
		\vspace{6pt}
		\STATE \textbf{def} $(A \boxtimes B)$:
		\COMMENT{$A$ and $B$ are of shape $N \times d$}
		\begin{ALC@g}
			\STATE $C \gets A$.reshape$(N \times d \times 1) \odot B$.reshape$(N \times 1 \times d)$
			\COMMENT{$\odot$ is the broadcasted Hadamard product. $C$ has shape $N \times d \times d$.}
			\STATE \textbf{return} $C$.reshape$(N \times d^2)$
			\vspace{6pt}
		\end{ALC@g}
		\STATE $\alpha \gets \sqrt[4]{d}$
		\STATE $V \gets \frac 1 N \left( \left( \sqrt{\frac{d}{N}}\1_N \right) \circ V \right) \in \R^{N \times d+1}$
		\STATE $Q, K \gets \frac{\alpha \tau Q}{\norm Q_{2, \text{dim}=-1}}, \frac{\alpha K}{\norm K_{2, \text{dim}=-1}}$
		\STATE $A_\text{mod} \gets (K \boxtimes K)^\top V$
		\STATE $\hat Y \gets (Q \boxtimes Q) A_\text{mod}$
		\STATE $\hat Y \gets \frac 1 2 \hat Y + \alpha^2 Q(K^\top V) + \alpha^4 \sum_{i = 1}^N V_{i, :}$
		\STATE $Y_\text{denom}, Y \gets \hat{Y}_{:, :1}, \hat{Y}_{:, 1:}$
		\STATE $Y \gets Y \oslash Y_\text{denom}$
		\COMMENT{$\oslash$ is the Hadamard division.}
		\RETURN $Y$
	\end{algorithmic}
\end{algorithm}

\subsection{Normalization}
\label{sec:normalization}
Empirical evaluations reveal the presence of intermediate values with large norms, which ultimately leads to failure to converge during training\footnote{See
\Cref{sec:appendix_no_normalization}
for further details.}.
Tracking the scaling behaviors (\Cref{tbl:scaling_behavior}) of intermediate results in \name\footnote{For more details see %
\Cref{sec:appendix_scaling_behavior}.} 
lets us define a normalization scheme that keeps these results from growing uncontrollably.

We first normalize the queries and keys and additionally introduce a per-head temperature parameter $\tau \in \R$\footnote{More details on the effect of normalizing compared to dividing by $d^{- \frac 1 2}$ (standard softmax attention does this) in %
\Cref{sec:appendix_const_norm_fact_d}.},
which ensures a constant input size:
\begin{align*}
	q_i &\gets \frac{\tau q_i}{\norm{q_i}_2}, \hspace{15pt} k_i \gets \frac{k_i}{\norm{k_i}_2}
	& \text{for } i = 1, ..., N.
\end{align*}
Then, we counteract the scaling behaviors in \Cref{tbl:scaling_behavior} by multiplying $Q$ and $K$ by $\sqrt[4]{d}$ and $V$ by $\frac{1}{N}$.
To obtain the same output, we need to scale the factors of the Taylor series accordingly\footnote{From $\frac 1 2, 1, 1$ to $\frac 1 2, \sqrt d, d$ ($\frac 1 2, \alpha^2, \alpha^4$ in Line 9 of \Cref{alg:taylored_attention}), to counteract the factors of $\sqrt[4] d$}.
To ensure a consistent mean size of the output $Y$ of \name, independent of $N$ and $d$, we additionally multiply by 
$\sqrt{\frac{N}{d}}$\footnote{To save on computations, we scale the denominator by $\sqrt{\frac{d}{N}}$ in Line 5 of \Cref{alg:taylored_attention}.}.
We add the same normalization of the input and output to direct-\name to keep both implementations interchangeable.
\Cref{alg:taylored_attention} shows the full procedure to calculate efficient-\name with normalization.

\section{Analysis of Efficiency Transition Points}
\label{sec:theoretical_analysis}
We have seen that efficient-\name has a complexity of $\O(Nd^3)$, while its direct version stands at $\O(N^2d)$.
Therefore, the efficient implementation will be faster and more memory efficient for sufficiently large sequence lengths $N \gg d$.
However, determining the exact value of $N$ where this transition occurs is crucial for practical scenarios.
This section analyzes the theoretical speed characteristics and memory requirements of both implementations to identify the specific point at which one outperforms the other independent of hardware considerations.
Furthermore, we analyze additional factors influencing the efficiency of both implementations, providing a deeper understanding of their performance.

\subsection{On the Floating-Point Operations}
\label{sec:on_flops}
To identify the critical sequence length $N_0$ at which the efficient implementation surpasses the direct one in a hardware- and implementation-agnostic way, we inspect the number of floating-point operations involved.
Starting with direct-\name, we follow \Cref{eq:taylor-attention} step by step.
We need $2N^2d$ operations to multiply $Q K^\top$, $4N^2$ operations to apply $x \mapsto \frac 1 2 x^2 + x + 1$ element-wise to this $N \times N$ matrix, $2N^2$ operations for normalization, and $2N^2d$ operations for the final multiplication by $V$.
The total FLOPS of direct-\name thus are
\begin{align} \begin{split} \label{eq:flops_triv_impl}
		\ops_\text{triv}[Y] &= 2N^2d + 4N^2 + 2N^2 + 2N^2d = 4N^2d + 6N^2.
\end{split}\end{align}
As the only difference between direct-\name and the standard attention mechanism is the choice of $\exp$ or its Taylor approximation, the number of operations needed for calculation of standard attention is slightly higher.

In contrast, for efficient-\name (\Cref{eq:efficient-implementation}), the primary computation centers around the squared influence $Y_\text{squ}$.
For $A_\text{mod} \in \R^{d^2 \times (d+1)}$ (\Cref{eq:a_mod_def}) the tensor operation has $N d^2$ FLOPS and the subsequent matrix multiplication needs $2 N d^2 (d+1)$. 
Factoring in the operations for the tensor operation on $Q$ and the second matrix multiplication, the total FLOPS for calculating $Y_\text{squ}$ are
\begin{align*}
	\ops[Y_\text{squ}] = 4Nd^2(d + 1) + 2Nd^2.
\end{align*}
Given the $4Nd(d + 1)$ operations required to compute the linear influence $Q K^\top V$, the $N (d+1)$ for summing up the columns of $V$, and the $3N(d+1)$ FLOPS for the sums and scalar multiplication, the total for calculating $\hat Y$ is
\begin{align*}
	\ops_\text{eff}[\hat Y] =& \ops[Y_\text{squ}] + \ops[Q K^\top V] + \ops[\Sigma_\text{col} V] + 3N(d + 1) \\
	=& 4 N d^2 (d+1) + 2 N d^2 + 4 N d (d + 1) + N(d + 1) + 3 N (d + 1).
\end{align*}
Including the $Nd$ operations for normalization, the total number of operations for efficient-\name is
\begin{align} \label{eq:flops_efficient}
	\ops_\text{eff}[Y] = N (4 d^3 + 10 d^2 + 9 d + 4).
\end{align}

\begin{table}[t]
	\centering
	\caption{Influence of the hidden dimension $d$ on the transitional points $N_0$ (speed) and $N_1$ (memory) based on \Cref{eq:N_0,eq:N_1} for typical $d$.}
	\begin{tabular}{l@{\hskip 4pt}@{\hskip 4pt}*4{r@{\hskip 8pt}}r}
		\toprule
		$d$   & 8 & 16 & 32 & 64 & 128 \\
		\cmidrule(r){1-1} \cmidrule{2-6}
		$N_0$ & 73 & 273 & 1057 & 4161 & 16513 \\
		$N_1$ & 47 & 159 & 574 & 2174 & 8446 \\
		\bottomrule
	\end{tabular}
	\label{tbl:N_0-values}
\end{table}

Comparing \Cref{eq:flops_triv_impl,eq:flops_efficient} shows that for ${N \to \infty}$, efficient-\name outperforms direct-\name, but for $N \not \gg d$ the latter will still be faster.
Let $N_0 = N_0(d)$ be the critical point, where $\ops_\text{triv}[Y] = \ops_\text{eff}[Y]$.
We calculate
\begin{align}
	N_0 &= \frac{4 d^3 + 10 d^2 + 9 d + 4}{4d+6} \leq d^2 + d + \frac 3 4. \label{eq:N_0}
\end{align}
For details on the derivation of $N_0$, see %
\Cref{sec:simplification_of_N0}.
Since the value of $d$ is typically fixed, we can easily compute the transitional input length $N_0$ for common choices of $d$.
The values for typical $d$ can be found in \Cref{tbl:N_0-values}.

\subsection{On Memory}
In addition to the number of operations, the memory footprint plays an important role as excessive memory needs can result in the inability to run a model altogether.
To assess it, we examine the largest tensors that have to be stored simultaneously, omitting memory needed for model parameters.

For direct-\name, maximum memory usage occurs when calculating the attention matrix $\taylorsm \left( Q K^\top \right)$ from $Q K^\top$.
Here, we store matrices $Q K^\top$ and $V$, as well as space for the output\footnote{Calculating the sum in $\frac{1}{2} x^2 + x$ requires saving the original value.} resulting in a total of
\begin{align*}
	\entr_\text{triv}[Y] = \underbrace{d N}_{\text{for } V} + \underbrace{2 N^2}_{\text{for } QK^\top \text{ and result}}.
\end{align*}

Conversely, the efficient version requires maximum memory during the calculation of $A_\text{mod}$ (\Cref{eq:a_mod_def}). 
Here, the matrices $(K^{\boxtimes 2})^\top$, $V$, and space for the result are needed, along with $Q$ and $K$ for later calculations for a total of
\begin{align}
	\entr_\text{eff}[Y] =& \underbrace{d^2 (d+1)}_{\text{for } A_\text{mod}} + \underbrace{2 d N}_{\text{for } Q, K} + \underbrace{(d + 1) N}_{\text{for } V} + \underbrace{d^2 N}_{\text{for } K^{\boxtimes 2}} \label{eq:mem_efficient}
\end{align}
matrix entries.
It is evident that
$\entr_\text{triv}[Y] > \entr_\text{eff}[Y]$
for all $N$ bigger than some constant $N_1 = N_1(d)$.
This marks the transitional point beyond which efficient-\name becomes more memory efficient than direct-\name.
By setting $\entr_\text{triv}[Y] = \entr_\text{eff}[Y]$ for $N = N_1$, we find
\begin{align}
	N_1 &= \frac 1 4 \left[ d^2 + 2 d + 1 + \sqrt{d^4 + 12 d^3 + 14 d^2 + 4 d + 1} \right] \leq \frac 1 2 d^2 + 2 d + \frac 1 2. \label{eq:N_1}
\end{align}
Refer to %
\Cref{sec:derivation_of_N1} 
for a detailed derivation.
Notably, from \Cref{tbl:N_0-values}, we observe that $N_1$ is considerably smaller than $N_0$ highlighting the extra memory efficiency of efficient-\name.

\subsection{Changing the Number of Attention Heads $h$}
\label{sec:changing_h}
In an effort to reduce the number of operations while retaining the ability to process the same number of tokens $N$, one might opt to reduce the internal dimension $d$.
However, this might come at the cost of expressiveness. 
Given that efficient-\name has a cubed complexity in $d$, an alternative strategy involves increasing the number of attention heads in the multi-head-attention mechanism. 
Let each token be $d_\text{emb} \in \N$ dimensional and let $h \in \N$ be the number of attention heads (with $h | d_\text{emb}$). 
Then, in each head, the queries, keys, and values are $d = \frac{d_\text{emb}}{h}$-dimensional, with the computational cost of the multi-head self-attention ($\text{MHSA}$) mechanism being $h$ times that of a single attention head. 
For direct-\name (\Cref{eq:flops_triv_impl}), the cost becomes
\begin{align*}
	\ops_\text{triv}[\text{MHSA}] &= h \ops_\text{triv}[Y] = h (4N^2d + 6N^2) = 4 N^2 d_\text{emb} + 6 h N^2,
\end{align*}
which strictly increases in $h$.
In contrast, using efficient-\name, we obtain
\begin{align*}
	\ops_\text{eff}[\text{MHSA}] &= h \ops_\text{eff}[Y] = h N (4 d^3 + 10 d^2 + 9 d + 4) \\
	&= N \left( 4 \frac{d_\text{emb}^3}{h^2} + 10 \frac{d_\text{emb}^2}{h} + 9 d_\text{emb} + 4h \right).
\end{align*}
Given that
$\ops_\text{eff}[\text{MHSA}]$ diverges for $h \to 0, \infty$,
there exists an optimal $\hat h_0 = \hat h_0(d_\text{emb})$ that minimizes the number of operations. %
Setting the derivative of $\ops_\text{eff}[\text{MHSA}]$ with respect to $h$ to zero, we find
\begin{align}
	0 &= \frac{\partial}{\partial h} \ops_\text{eff}[\text{MHSA}] = N \left( 4 - 9 \frac{d_\text{emb}^3}{h^3} - 10 \frac{d_\text{emb}^2}{h^2} \right)
	\overset{N > 0}{{\Leftrightarrow}} 4 = 9 d^3 + 10 d^2.
	\label{eq:d_def}
\end{align}
This has a single positive solution of
$d \approx 0.52$, minimizing the number of operations at
$\hat h_0 \approx \frac{1}{0.52} d_\text{emb}$.
For a detailed derivation refer to %
\Cref{sec:derivation_of_z}.
In particular, the number of operations of efficient-\name decreases when $h$ increases in the range of possible values $\{1, 2, ..., d_\text{emb} \}$ (divisors of $d_\text{emb}$).

Examining memory costs provides another perspective on the impact of attention heads.
On one hand, for direct-\name the number of simultaneous entries strictly increases with the number of attention heads $h$, when calculating heads in parallel:
\begin{align*}
	\entr_\text{triv}[\text{MHSA}] = h \entr_\text{triv}[Y] = d_\text{emb} N + 2 N^2 h
\end{align*}
On the other, for efficient-\name, the number of entries is
\begin{align*}
	\entr_\text{eff}&[\text{MHSA}] = h \entr_\text{eff}[Y] = h (d^3 + (N+1) d^2 + 3 Nd + N) \\
	&= \frac{d_\text{emb}^3}{h^2} + (N + 1) \frac{d_\text{emb}^2}{h} + 3N d_\text{emb} + Nh.
\end{align*}
This expression again diverges as $h \to 0, \infty$ and therefore an optimum $\hat h_1$ exists. %
Setting the derivative to zero gives
\begin{align}
	0 &= \frac{\partial}{\partial h} \entr_\text{eff}[\text{MHSA}] = -2 d^3 - (N + 1) d^2 + N,
\end{align}
which implies $d < 1$ and therefore
$\hat h_1 > d_\text{emb}$.
Refer to %
\Cref{sec:appendix_derivation_of_h1} 
for the detailed derivation.
In particular, the memory cost also decreases with increasing $h$ in the allowed range $\{1, ..., d_\text{emb}\}$.
The same holds true when calculating the attention heads in sequence (\Cref{eq:mem_efficient} is strictly increasing in $d$).
Our analysis provides insight into the dynamic efficiency interplay between the two implementations and the number of attention heads.

\begin{figure*}[t]
	\centering
	\includegraphics[width=.9\textwidth]{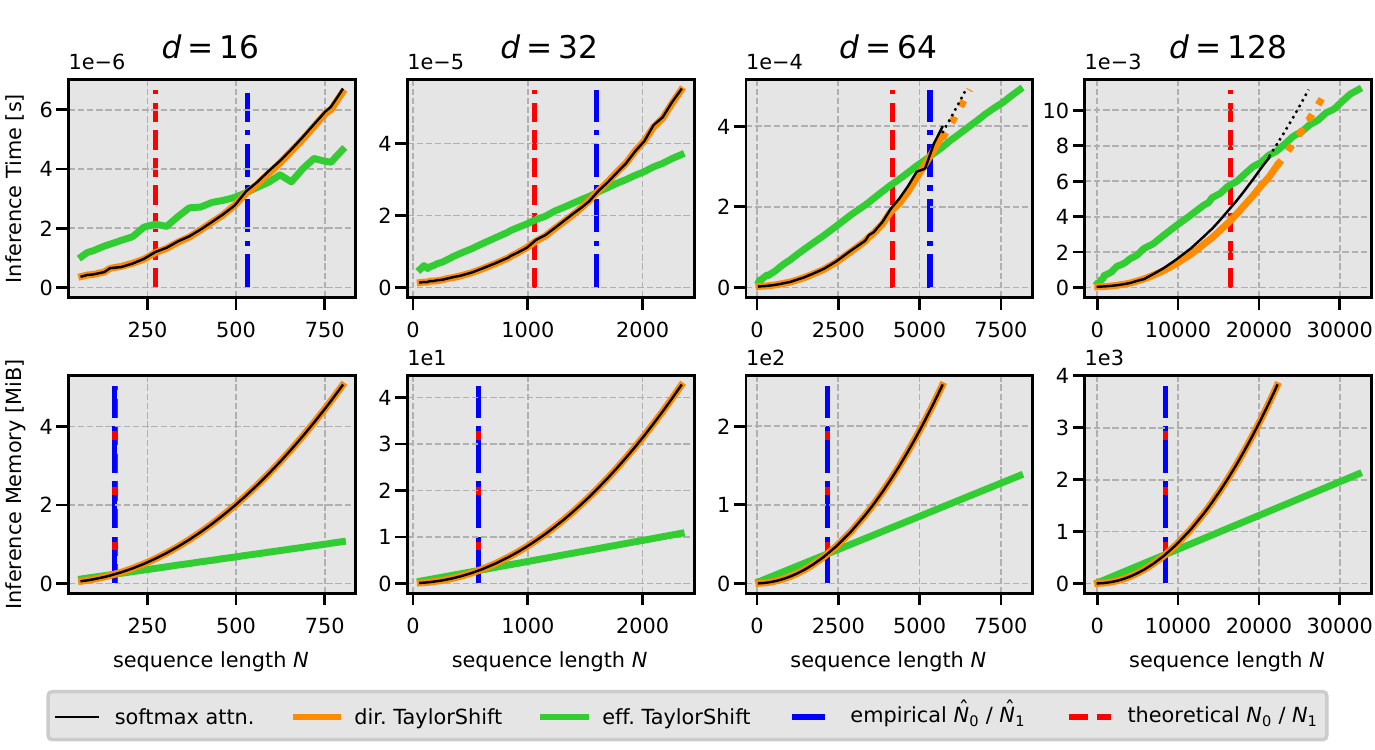}
	\caption{Inference time in seconds per input (top) and inference memory in MiB (bottom) of the attention mechanism (with $h = 1$) vs. sequence length for both implementations of \name and softmax attention. Each column uses a different internal dimension $d$. We mark the theoretical $N_0$ and $N_1$ and empirical intersections $\hat N_0$ and $\hat N_1$. Dotted lines extrapolate values by fitting a parabola.}
	\label{fig:empirical_cutoffs}
\end{figure*}

\section{Empirical Evaluation}
\label{sec:experiments}
We run a number of experiments that provide an empirical verification of our theoretical analysis of the transitional bounds, scalability, and required computational resources, as well as of the effective capacity of our proposed mechanism.

\subsection{Efficiency of the \name module}

To validate our theoretical analysis of the critical points $N_0$ and $N_1$ from \Cref{sec:theoretical_analysis}, we compare the speed and memory usage of \name and softmax attention \cite{Vaswani2017} using simulated data.
For multiple internal dimensions $d$ and sequence lengths $N$, we measure inference time and memory consumption of a single attention head on an NVIDIA A100 GPU.
For comparison, applications like GPT-2 \cite{Radford2019} or ViT \cite{Dosovitskiy2021}, use a per-head dimension of $d = 64$.

In \Cref{fig:empirical_cutoffs} (top), we contrast the speed of \name and softmax attention. 
The quadratic growth of softmax attention and direct-\name and the linear growth of efficient-\name are evident.
As noted in \Cref{sec:on_flops}, we observe a slightly higher number of FLOPS for softmax attention than for direct-\name.
Note that the difference between the theoretical $N_0$ and empirical $\hat N_0$ transition points $\hat N_0 - N_0 \approx 18 d$ is approximately proportional to 
$d$.
We hypothesize that the more sequential nature of efficient-\name results in more, costly reads and writes in GPU memory. This indicates possible efficiency gains for eff. \name from a low-level IO-efficient implementation.\footnote{For more details, see %
\Cref{sec:emp_theo_diff_details}.}

Due to increasing memory requirements for direct-\name and softmax attention, plotted in the second row, we need to extrapolate the plots for $d = 64$ and $d = 128$ by fitting a parabola (dotted lines) to the data
In the regimen of memory (second row of \Cref{fig:empirical_cutoffs}), the theoretical and empirical intersections align closely $\hat N_1 \approx N_1$, with an error of at most $0.6\%$.
Comparing both rows shows efficient-\name becoming memory efficient earlier than it becomes efficient in terms of speed, highlighting its usefulness in low-memory environments, in alignment with our theoretical results from \Cref{tbl:N_0-values}.

\begin{figure}[t]
	\centering
	\includegraphics{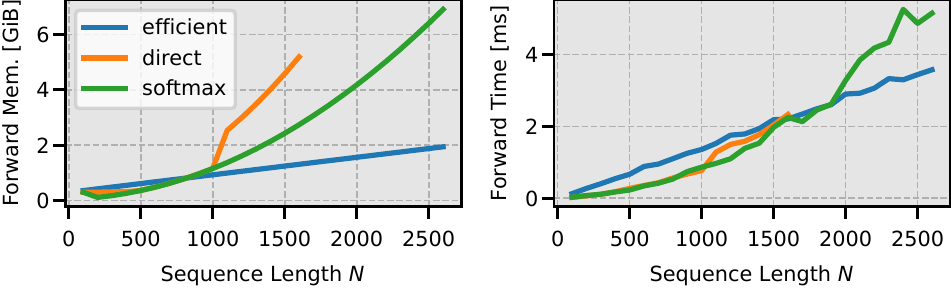}
	\caption{Memory and inference time of a transformer with efficient- and direct-\name and the standard softmax, using $d=32$.
	}
	\label{fig:real_world_mem_and_tp}
\end{figure}
\subsection{Efficiency of a Transformer with \name}
We show the efficiency of a full-scale\footnote{Here, we use the hyperparameters for ListOps from %
\Cref{sec:apendix_experimental_setup}, 
but with 16 heads.} Transformer encoder equipped with \name in \Cref{fig:real_world_mem_and_tp}.\footnote{The extended \Cref{fig:real_world_mem_and_tp} in %
\Cref{sec:appendix_eff_comparison} 
includes different numbers of heads.}
At a sequence length of 900 tokens efficient-\name needs less memory and at 1800 tokens it surpasses the standard Transformer in speed.
Note that at 1500 tokens it only needs half and at 2000 tokens only 35\% of the Transformer's memory.
For shorter sequence length, direct-\name remains competitive with a standard Transformer in terms of speed and memory.

\subsection{Performance of a Transformer with \name}
\begin{table*}[t]
	\centering
	\caption{Accuracy in percent for models on datasets of different modalities.
		For the first three datasets, we closely adhere to the setup of \cite{Tay2021a}.
		Models with ${}^\star$ had to be trained with full instead of mixed precision.}
	\label{tab:performance_comparison}
	\resizebox{.9\textwidth}{!}{
	\begin{tabular}{l*5{S[table-format=2.1]}c}
		\toprule
		\multirow{2}{*}{Model}                                                             & {CIFAR}          & {IMDB}           & {ListOps}        & {ImageNet}       & {ImageNet}       &    {Average}     \\
		                                                                                   & {(Pixel)}        & {(Byte)}         &                  & {(Ti)}           & {(S)}            &  \\
		\cmidrule(r){1-1} \cmidrule(lr){2-6} \cmidrule(l){7-7}
		Linformer \cite{Wang2020} 														   & 29.2             & 58.1             & {-}              & 64.3             & 76.3             &     {(57.0)}     \\
		RFA	\cite{Peng2021a}                                                               & 44.9             & \underline{65.8} & {-}              & {-}              & {-}              &      (55.4)      \\
		Performer	\cite{Choromanski2021}                                                 & 34.2${}^\star$   & 65.6${}^\star$   & 35.4${}^\star$   & 62.0${}^\star$   & 67.1${}^\star$   &       52.9       \\
		Reformer \cite{Kitaev2020}                                                         & 44.8             & 63.9             & \textbf{47.6}    & 73.6             & 76.2$^\star$     &       61.2       \\
		Nystromformer \cite{Xiong2021}                                                     & \textbf{49.4}    & 65.6             & 44.5             & \underline{75.0} & 78.3${}^\star$   & \underline{62.6} \\
		EVA	\cite{Zheng2023}                                                               & 46.1             & 64.0             & 45.3             & 73.4             & 78.2             &       61.4       \\
		Transformer \cite{Vaswani2017}                                                     & 44.7             & \underline{65.8} & 46.0             & \textbf{75.6}    & \underline{79.1} &       62.2       \\
		\textbf{Ours}                                                                      & \underline{47.6} & \textbf{66.2}    & \underline{46.1} & \underline{75.0} & \textbf{79.3}    &  \textbf{62.8}   \\ %
		\bottomrule
	\end{tabular}}
\end{table*}
To assess the effectiveness of \name, we evaluate it using a Transformer-encoder across various datasets representing different modalities.
We track the classification accuracy of a \name-equipped Transformer across five tasks.

\subsubsection{Tasks}
We train on three datasets introduced by {\cite{Tay2021a}}, specially designed to assess performance on long sequences with long-range dependencies.
The first is a pixel-level CIFAR10 task, where 8-bit intensity values of grayscale images from CIFAR10 are encoded into a sequence 
of length 1024.
In the domain of text, IMDB Byte \cite{Maas2011}, is a classification task for text encoded at the character level, resulting in sequences of 4000 tokens. %
Thirdly, we employ the Long ListOps dataset of mathematical operations \cite{Nangia2018} of length 500 to 2000 tokens encoded at the character level.
Beyond these synthetic tasks, we train for classification on ImageNet \cite{Deng2009}
at two sizes (Ti \& S) to additionally evaluate the scaling behavior of \name.
Refer to %
\Cref{sec:apendix_experimental_setup} 
for model sizes and training hyperparameters.
We utilize mixed-precision calculations whenever possible.

\Cref{tab:performance_comparison} shows our method's consistent performance across all datasets.
It surpasses all other linear scaling Transformers on a minimum of four out of five datasets.
Note, that those models marked with ${}^\star$ only work using full precision, slowing down training considerably.
\name also outperforms the standard Transformer on four out of five tasks, remains competitive on the last one.
We observe a notable increase of $4.3\%$ when transitioning from size Ti to S on ImageNet, in contrast to $3.5\%$ for the Transformer.
These findings highlight the robustness and competitiveness of \name across diverse datasets and modalities.
This demonstrates \name's usefulness when dealing with very long sequences.

\subsection{Ablations}
We conduct an ablation analysis, systematically dissecting two key components to establish their impact on the performance of \name.
\paragraph{Normalization}
\begin{table}[t]
	\centering
	\caption{Accuracy on the CIFAR Pixel task when ablating our novel normalization introduced in \Cref{sec:normalization}.}
	\label{tab:ablations_normalization}
	\begin{tabular}{lcc}
		\toprule
		Model & direct & efficient \\
		\cmidrule(r){1-1} \cmidrule(l){2-3}
		Plain impl. & 47.1 & - \\
		impl. +norm. & 46.8 & 46.8 \\
		impl. +norm. +output norm. & 47.5 & 47.6 \\ 
		\bottomrule
	\end{tabular}
\end{table}
We train a Transformer equipped with \name at different stages of normalization to track the impact of our normalization scheme.
\Cref{tab:ablations_normalization} shows that without normalization, direct-\name demonstrates acceptable performance, while the efficient version fails to converge during training.
We attribute this to numerical overflow in intermediate results\footnote{See also %
\Cref{sec:appendix_no_normalization}.}.
Upon introducing input normalization to the attention mechanism, efficient-\name becomes stable, and both implementations achieve an accuracy of 46.8\%, a slight decrease for direct-\name.
Additionally, normalizing the output to a mean size of 1, results in a performance boost for both implementations, bringing them to the accuracy level observed for the direct version before.

\paragraph{Number of attention heads $h$}
\begin{table}[t]
	\centering
	\caption{Accuracy, throughput (TB), and VRAM (Mem) usage of \name on the CIFAR Pixel task with different number of attention heads $h$.
		All models have $d_\text{embed} = 256$ with $d = \frac{d_\text{embed}}{h}$ in the attention mechanism.
	}
	\label{tab:ablations_heads}
	\small
	\begin{tabular}{S[table-format=2]S[table-format=2]c@{\hskip 6pt}r@{\hskip 6pt}r@{\hskip 6pt}r@{\hskip 6pt}r}
		\toprule
		{\multirow{2}{*}{{$h$}}} & {\multirow{2}{*}{{$d$}}} & {\multirow{2}{*}{{Acc [\%]}}} & \multicolumn{2}{c}{direct} & \multicolumn{2}{c}{efficient} \\
		&  & &  {TP [ims/s]} & Mem [MiB@16] & TP [ims/s] & Mem [MiB@16] \\
		\cmidrule(r){1-2} \cmidrule(lr){3-3} \cmidrule(lr){4-5} \cmidrule(l){6-7}
		4 & 64 & 47.1 & 12\,060 & 596 & 2\,975 & 840 \\
		8 & 32 & 47.5 & 7\,657 & 1\,111 & 5\,749 & 585 \\
		16 & 16 & 47.3 & 4\,341 & 2\,135 & 9\,713 & 459 \\
		32 & 8 & 46.9 & 2\,528 & 4\,187 & 14\,087 & 397 \\
		64 & 5 &45.9 & 1\,235 & 8\,291 & 13\,480 & 125 \\
		\bottomrule
	\end{tabular}
\end{table}
Finally, we validate our insights from \Cref{sec:changing_h} by training a \name-equipped encoder with varying numbers of attention heads $h$ while maintaining the embedding dimension $d_\text{embed}$.
Note that the number of parameters stays almost exactly constant, with only the shape of the attention temperature per head ($\tau$) changing.
The results in \Cref{tab:ablations_heads} align with our theoretical analysis, demonstrating an acceleration and less memory demands as the number of heads increases.
Notably, increasing the number of heads often leads to increased accuracy while concurrently speeding up calculations and reducing memory.
These efficiency gains will become more significant for sequences longer than the 1024 tokens we tested with.
Beyond the point where accuracy increases, we can still leverage additional heads to trade off accuracy against speed and memory, particularly advantageous for processing longer sequences.

\section{Conclusion}
\label{sec:conclusion}

We present \name, an efficient attention mechanism that computes token-to-token interactions in linear time and memory.
We lay the theoretical groundwork for using \name by studying the exact threshold values where it becomes efficient.
Empirical validation of our analysis through classification experiments confirms the performance benefits of \name for long sequences.
\name even outperforms a standard Transformer across diverse datasets and modalities by 0.6\% on average, while being faster and using less than half the memory for sequences longer than 2000 tokens.
Furthermore, our results on the number of attention heads reaffirm the efficiency gains predicted theoretically.
The number of heads can be tuned to improve the model's effective capacity, its speed, and reduce memory requirements, all at once.
While efficient-\name is faster than a standard Transformer for long sequences, we can swap back to the interchangeable direct-\name variant to keep the model efficient for short sequences.
This can be useful when dealing with datasets containing sequences of vastly different length, like text or time-series, or when using a curriculum to build up to very long sequence tasks.
By adopting \name, it will be possible to tackle tasks featuring long sequences such as high-resolution image classification and segmentation, processing long documents, integrating data from multiple modalities, and dynamically encoding lengthy documents into a prompt-specific context for Large Language Models.
Overall, our findings underscore the efficiency and versatility of \name, positioning it as a competitive and scalable option in the landscape of efficient attention-based models.

\subsubsection{Acknowledgments}
This work was funded by the Carl-Zeiss Foundation under the Sustainable Embedded AI project (P2021-02-009).

 \bibliographystyle{splncs04}
 \bibliography{TaylorShift_ICPR}
\newpage

\appendix

\section{Mathematical Details}
\subsection{Derivation of $N_0$}
\label{sec:simplification_of_N0}

Simplification of  $N_0$:
\begin{align*}
	N_0 &= \frac{4d^3 + 10 d^2 + 9d + 4}{4d + 6} \\
	&= \frac{4 d^3 + 6 d^2}{4d + 6} + \frac{4d^2 + 6d}{4d + 6} + \frac{3d + 4}{4d + 6} \\
	&\leq \frac{4 d^3 + 6 d^2}{4d + 6} + \frac{4d^2 + 6d}{4d + 6} + \frac{3d + 4.5}{4d + 6} \\
	&= d^2 + d + \frac 3 4
\end{align*}

\subsection{Derivation of $\hat h_0$}
\label{sec:derivation_of_z}

To find $\hat h_0$, we want to find $d = \frac{d_\text{embed}}{h} \in \R$, such that 
\begin{align}
	9 d^3 + 10 d^2 = 4
\end{align}
holds.
\begin{align*}
	9 d^3 + 10 d^2 &= 4 \\
	\xLeftrightarrow{d = x - \frac{10}{27}} 9 x^3 - \frac{100}{27} x - \frac{6748}{2187} &= 0 \\ 
	\Leftrightarrow x^3 - \frac{100}{243} x - \frac{6748}{19683} &= 0 \\
	\xLeftrightarrow{x = y + \frac{100}{729 y}} y^3 - \frac{6748}{19683} + \frac{1 000 000}{387 420 489} y^{-3} &= 0 \\
	\xLeftrightarrow{u = y^3 | \cdot u} u^2 - \frac{6748}{19683} u + \frac{1 000 000}{387 420 489} &= 0.
\end{align*}

The last equation has the solution
\begin{align*}
	u = \frac{3374 + 54 \sqrt{3561}}{19683}.
\end{align*}
Then we can substitute $\alpha := \sqrt[3]{3374 + 54 \sqrt{3561}} \approx 18.75$ and $u = y^3$ using $\zeta_3 = e^{\frac 2 3 i \pi}$ a third root of unity, to get
\begin{align*}
	y &= \frac{\zeta_3^j}{27} \alpha \\
	\Rightarrow x &=  y + \frac{100}{729 y} = \frac{\zeta_3^j}{27} \alpha + \frac{100}{729} \alpha^{-1} \zeta_3^{-j} \\
	\Rightarrow d &= x - \frac{10}{27} = \frac{\zeta_3^j}{27} \alpha + \frac{100}{729} \alpha^{-1} \zeta_3^{-j} - \frac{10}{27}
\end{align*}
for $j = 0, 1, 2$.

Now, by the property of the third roots of unity, we have $\Im \zeta_3^j = - \Im \zeta_3^{-j}$. Since $\alpha \neq \frac{100}{27 \alpha}$, $d$ is real if and only if $j = 0$. Therefore, the only real solution to Equation (11) of the main paper
is
\begin{align*}
	d = \frac{1}{27} \alpha + \frac{100}{729} \alpha^{-1} - \frac{10}{27} \approx 0.52.
\end{align*}

\subsection{Derivation of $\hat h_1$}
\label{sec:appendix_derivation_of_h1}
The goal is to find the optimum number of attention heads which implicitly fulfills
\begin{align*}
	0 &= -2 d^3 - (N + 1) d^2 + N \\
	\Leftrightarrow N &= 2 d^3 + (N + 1) d^2 = (2d + N + 1) d^2 \overset{d > 0}{\geq} (N + 1) d^2.
\end{align*}
Therefore it holds
\begin{align*}
	1 > \frac{N}{N + 1} \geq d^2,
\end{align*}
which implies $1 > d = \frac{d_\text{embed}}{\hat h_1}$ and $\hat h_1 > d_\text{embed}$.

\subsection{Derivation of $N_1$}
\label{sec:derivation_of_N1}

We have $h d N_1 + 2 h N_1^2 = h d^2 (d+1) + 2 h d N_1 + h (d + 1) N_1 + h d^2 N_1$ by definition of $N_1$.
Therefore
\begin{align*}
	d^2 (d+1) + 2dN_1 + (d + 1) N_1 + d^2 N &= dN + 2N^2 \\
	\Leftrightarrow N^2 - \frac{d^2 + 2d + 1}{2} N - \frac{d^3 + d^2}{2} &= 0,
\end{align*}
which has two solutions. The larger of those being
\begin{align*}
	N_1 &= \frac{1}{4} \left[ d^2 + 2d + 1 + \sqrt{(d^2 + 2d + 1)^2 + 8 (d^3 + d^2)} \right] \\
	&= \frac 1 4 \left[ d^2 + 2d + 1 + \sqrt{d^4 + 12 d^3 + 14 d^2 + 4 d + 1} \right].
\end{align*}

Since
\begin{align*}
	(d^2 + 6 d + 1)^2 &= d^4 + 12 d^3 + 38 d^2 + 12 d + 1 \\
	&\geq d^4 + 12 d^3 + 14d^2 + 4d +1,
\end{align*}
we have
\begin{align*}
	N_1 \leq \frac{1}{2} d^2 + 2d + \frac 1 2.
\end{align*}

\section{Normalization \& Numerical Behavior}
\label{sec:appendix_normalization}
\subsection{Training Without Normalization}
\label{sec:appendix_no_normalization}
\begin{figure}[t]
	\centering
	\resizebox{.5\textwidth}{!}{\includegraphics{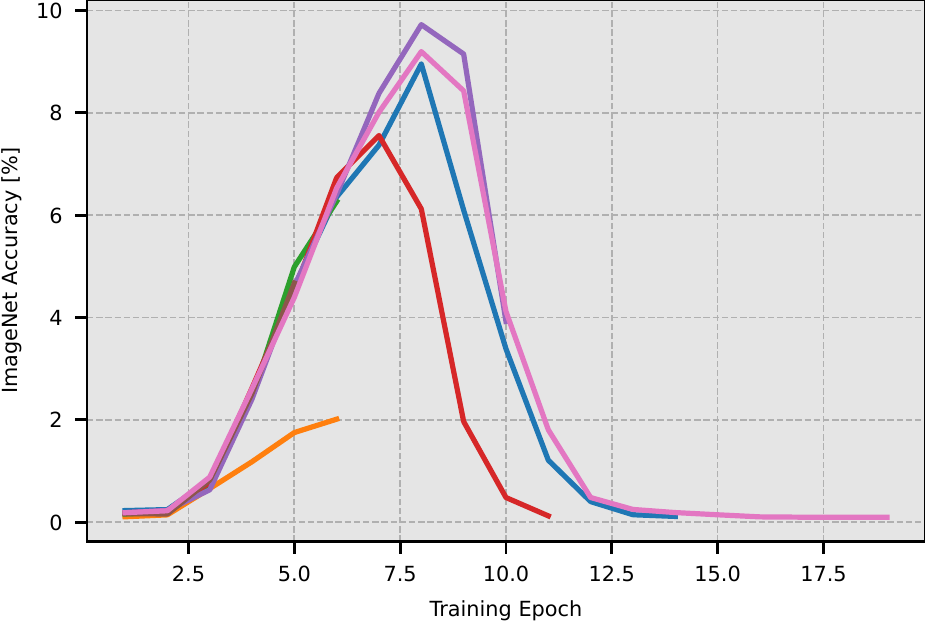}}
	\caption{ImageNet accuracy in early training when using the efficient implementation without normalization during validation. These models have been trained at a sequence length of only $N = 197$ using different hyperparameters.}
	\label{fig:non_normalized_runs}
\end{figure}
We find that not using normalization leads to numerical instabilities during training.
Large intermediate results quickly lead to degenerating performance due to numerical errors and training often breaks due to overflow-induced NaN-values.
\Cref{fig:non_normalized_runs} shows a few training runs, where normalization has been turned of at test time.
These curves first display the influence of numerical inaccuracies while stopping after only a hand full of epochs, as numerical overflows render further calculations impossible.
Our novel normalization scheme eliminates these types of training failures.

\subsection{Scaling Behavior}
\label{sec:appendix_scaling_behavior}
\begin{figure*}[t]
	\centering
	\includegraphics[width=\textwidth]{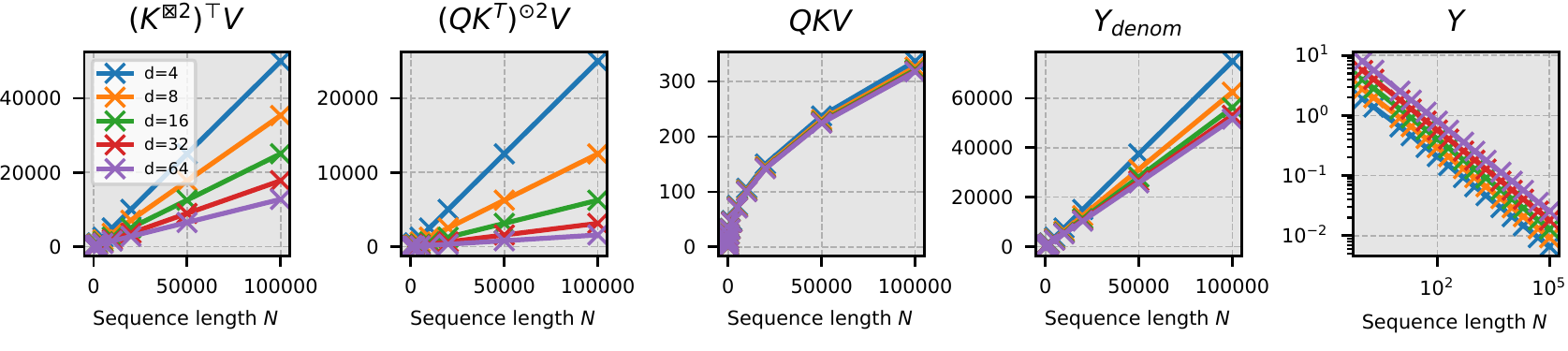}
	\caption{Mean norm of different expressions at sequence length $N$ from $1$ to $100 000$ with ${Q,K,V \sim \text{unif}\left( \mathcal  S^{d + 1} \right)}$. Calculated using $16 384$ samples each.}
	\label{fig:scaling_behavior}
\end{figure*}

\begin{figure*}[t]
	\centering
	\includegraphics[width=\textwidth]{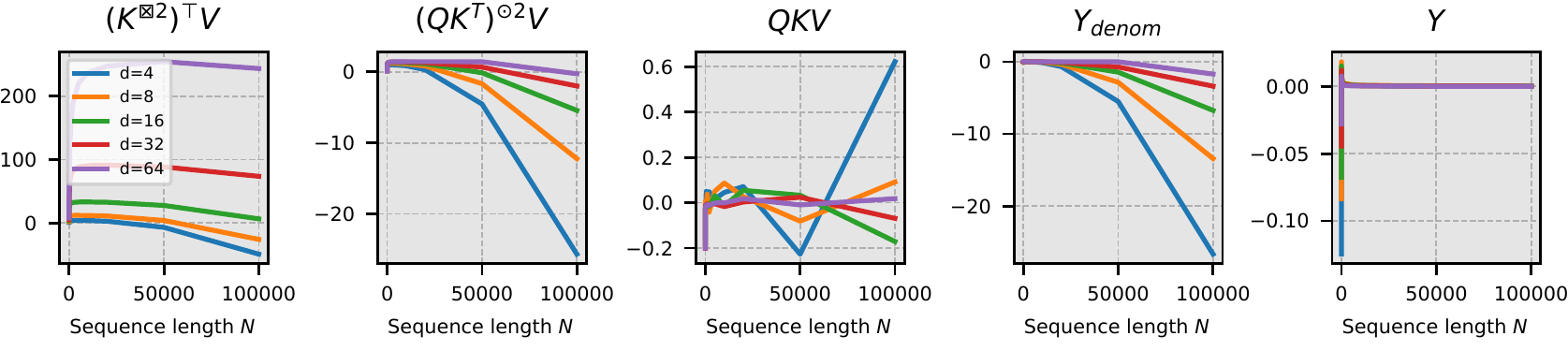}
	\caption{Absolute errors of our fitted scaling behaviors from Table 1 of the main paper
		to empirical values in \Cref{fig:scaling_behavior}.}
	\label{fig:scaling_behavior_errors}
\end{figure*}
To analyze the scaling behavior of \name and inform our normalization (see Section 3.3 
in the main paper), take a look at the size of intermediate results of our calculations when varying $d$ and $N$.
We uniformly sample the matrices $Q, K, V \in \R^{N \times d}$ from the unit sphere, as our formulation uses normalized queries and keys.
We then measure the mean vector norm of intermediate results.
Experimental results of the scaling behavior of intermediate expressions of our efficient implementation are shown in \Cref{fig:scaling_behavior}.
The specific formulas used for normalization (Table 1 of the main paper)
were derived based on simple candidate functions fitted to empirical results, considering factors such as growth behavior and critical point.
\Cref{fig:scaling_behavior_errors} shows the errors of in these fitted functions.
These errors are $\leq 1\%$ for large sequence length $N$, making our approximations useful for normalization.

\subsection{Normalization by $d^{- \frac 1 2}$ in Softmax Attention}
\label{sec:appendix_const_norm_fact_d}

We do not normalize by a factor of $\frac{1}{\sqrt d}$, since this would make no difference when done before normalizing the queries and keys.
The scaling factor of $\frac{1}{\sqrt d}$ was originally introduced in \cite{Vaswani2017} to avoid $QK^\top$ values growing too large, when the per-head dimension is large, to avoid vanishing gradients from the $\softmax$.
In our case, we normalize the query and key vectors, which has the same effect of preventing values of the $QK^\top$ matrix from growing too much.
Furthermore, our model utilizes a per-head attention temperature ($\tau$) to learn the optimal range of $QK^\top$ values during training, allowing it to adapt and adjust the scaling factors as needed.
In practice, we observe a wide range of scaling factors from $10$ to $80$ for TaylorShift trained on ImageNet (hyperparameters for the small version -- S).

\section{Experimental Setup}
\label{sec:apendix_experimental_setup}

\subsection{Hyperparameters}
\begin{table}[t]
	\centering
	\caption{Model sizes and training hyperparameters that were used for all models, depending on the dataset.
		These hyperparameters are based on \cite{Touvron2022} for ImageNet and on \cite{Tay2021a} for the other datasets.
		The lr schedule, warmup epochs, weight decay, dropout, and drop path rate were the same for all models.
		We trained on NVIDIA A100 GPUs.}
	\label{tab:training_hyperparameters}
	\resizebox{\textwidth}{!}{
		\begin{tabular}{rccccc}
			\toprule
			param & CIFAR (Pixel) & IMDB (Byte) & ListOps & ImageNet (Ti) & ImageNet (S) \\
			\cmidrule(r){1-1} \cmidrule(lr){2-4} \cmidrule(l){5-6}
			model depth & 1 & 4 & 4 & 12 & 12 \\
			$d_\text{embed}$ & 256 & 256 & 512 & 192 & 348 \\
			heads $h$ & 4 & 4 & 8 & 3 & 6 \\
			MLP ratio & 1 & 4 & 2 & 4 & 4\\
			\cmidrule(r){1-1}
			lr & 5e-4 & 5e-5 & 1e-3 & \multicolumn{2}{c}{3e-3} \\
			batch size & 256 & 32 & 256 & \multicolumn{2}{c}{2048} \\
			epochs & 200 & 200 & 200 & \multicolumn{2}{c}{300} \\
			lr schedule & \multicolumn{5}{c}{cosine decay} \\
			warmup epochs & \multicolumn{5}{c}{5} \\
			weight decay & \multicolumn{5}{c}{1e-3}\\
			pos. embed. & cosine & cosine & cosine & \multicolumn{2}{c}{learned} \\
			dropout & \multicolumn{5}{c}{0}\\
			drop path rate & \multicolumn{5}{c}{0.05}\\
			optimizer & \multicolumn{5}{c}{fused LAMB} \\
			mixed precision & \multicolumn{5}{c}{whenever possible} \\
			data augmentation & - & - & - & \multicolumn{2}{c}{3-augment \cite{Touvron2022}} \\
			GPUs & 4 & 4 & 8 & 4/8 & 8 \\
			\bottomrule
	\end{tabular}}
\end{table}
\Cref{tab:training_hyperparameters} shows the hyperparameters we used for training on the different datasets.
Our hyperparameter choices and model sizes are based on \cite{Tay2021a} for the CIFAR, IMDB, and ListOps datasets and on \cite{Touvron2022} for ImageNet.
For IMDB and CIFAR, we used Byte-level encoding. ListOps is encoded at the character level (17 possible characters), and for ImageNet, we encoded RGB-patches of size $16 \times 16$. 

\subsection{Baseline Models}
We compare \name against a handful of linear scaling efficient Transformers, starting with the Linformer \cite{Wang2020}, which projects down the sequence direction into a lower dimensional space.
Nyströmformer \cite{Xiong2021} utilizes a Nyström decomposition to approximate the attention matrix $A = \softmax \left( Q K^\top \right)$ in linear time.
We compare to the Kernel attention based methods RFA \cite{Peng2021a}, Performer \cite{Choromanski2021}, and EVA \cite{Zheng2023} which all approximate the exponential function using Gaussian random variables, adding different methods on top to improve the approximation.
These turn out to be the most similar to efficient \name, structurally.
We find that these models tend to be unstable during training, exemplified by RFA failing to converge on ListOps and ImageNet and Performer requiring full precision to converge.
Our normalization procedure alleviates those kinds of issues in efficient \name.
Additionally, we compare to Reformer \cite{Kitaev2020}, which implements sparse attention by clustering tokens.
Reformer has a complexity of $\O (N \log N)$ in the sequence length $N$.
Last and most importantly, we of course compare to a standard Transformer Encoder \cite{Vaswani2017}.
All models use the same set of standard hyperparameter values.

\subsection{Level of Implementation}
We chose to compare models at the algorithmic level rather than the implementational level, as our primary focus is to assess the intrinsic efficiency of different attention mechanisms, independent of specific optimization techniques or hardware dependencies.
To archive this and also be able to have a meaningful empirical comparison, we choose implementations for each algorithm at a similar level. While there certainly are implementations of self-attention that are more optimized, most notably Flash \cite{Dao2022}, these kinds of optimized implementations are not available for the efficient attention mechanisms, rendering it a biased comparison.
Since it is possible to speed up every attention mechanism by engineering an IO-aware implementation, we consider this route to be orthogonal to our contribution and out of the scope of this paper.

Instead, we implement every attention mechanism at a higher level of abstraction, using PyTorch \cite{Paszke2019}.
In particular, for the implementation of the standard attention mechanism, we fall back to an implementation from Timm \cite{Wightman2019}\footnote{v0.8.10: \url{https://github.com/huggingface/pytorch-image-models/blob/1e0b34722772b6612ceab18cfe43d2e6a10c204e/timm/models/vision_transformer.py\#L66}}.

\subsection{Datasets and Tasks}
We run experiments on a handful of tasks from the Long Range Arena Benchmark \cite{Tay2021a}.
These are specifically engineered to test the performance of non-causal self-attention mechanisms on very long sequences.

\paragraph{CIFAR10 Pixel}
The CIFAR Pixel task was designed to test the attention mechanisms ability to learn complex 2D-relationships from the 1D input sequence. The CIFAR10 images are transformed to 8-bit gray-scale. Then each pixel is turned into a token by individually encoding each 8-bit value.
The resulting sequence has 1024 tokens.

\paragraph{IMDB Byte}
This task goes into the domain of language processing. Text from the IMDB Dataset \cite{Maas2011} is encoded at the byte/character level, similar to the CIFAR task, to increase the sequence length and task difficulty.
The resulting sequences are cut/padded to length 4000 and then classified into two classes.

\paragraph{Long ListOps}
The task is to solve long mathematical operations. The sequences consist of mathematical operators $\min, \max, \operatorname{median}, \operatorname{first}, \operatorname{last},$ and $\operatorname{sum} \mod 10$ together with a sequence of digits and other operators to create nested sequences of depth $\leq 10$.
The result is modeled as a classification task on the 10 possible outputs and sequences are again encoded at the character level.
We procedurally generate batches with sequences of consistent length from 500 to 2000.

\subsubsection{Task Correlation}
CAB \cite{Zhang2023b} shows a very high correlation between LRA score and the performance of non-causal self-attention for other more realistic tasks using language and speech. This validates the suitability of the LRA benchmark for tasks utilizing non-causal self-attention. Additionally, CAB points out that the use of different attention variants (causal/non-causal self-/cross-attention) is not (positively) correlated, but our focus lies on the non-causal self-attention setting.

\paragraph{ImageNet}
In order to also evaluate performance on real-world data, we include the ImageNet \cite{Deng2009} classification task.
Here, we utilize the standard approach from \cite{Dosovitskiy2021} of cutting the image into patches of size $16 \times 16$ that are linearly embedded into tokens. Then, a learnable positional encoding is added.
While using the standard image size of $224 \times 224$ px only results in sequences of length 196, we use this task to evaluate the performance of attention mechanisms on complex real data.

\section{Further Analysis}
\subsection{Point of Taylor Expansion}
\label{sec:appendix_point_of_taylor_expansion}
\begin{figure}[t]
	\centering
	\resizebox{\textwidth}{!}{\includegraphics{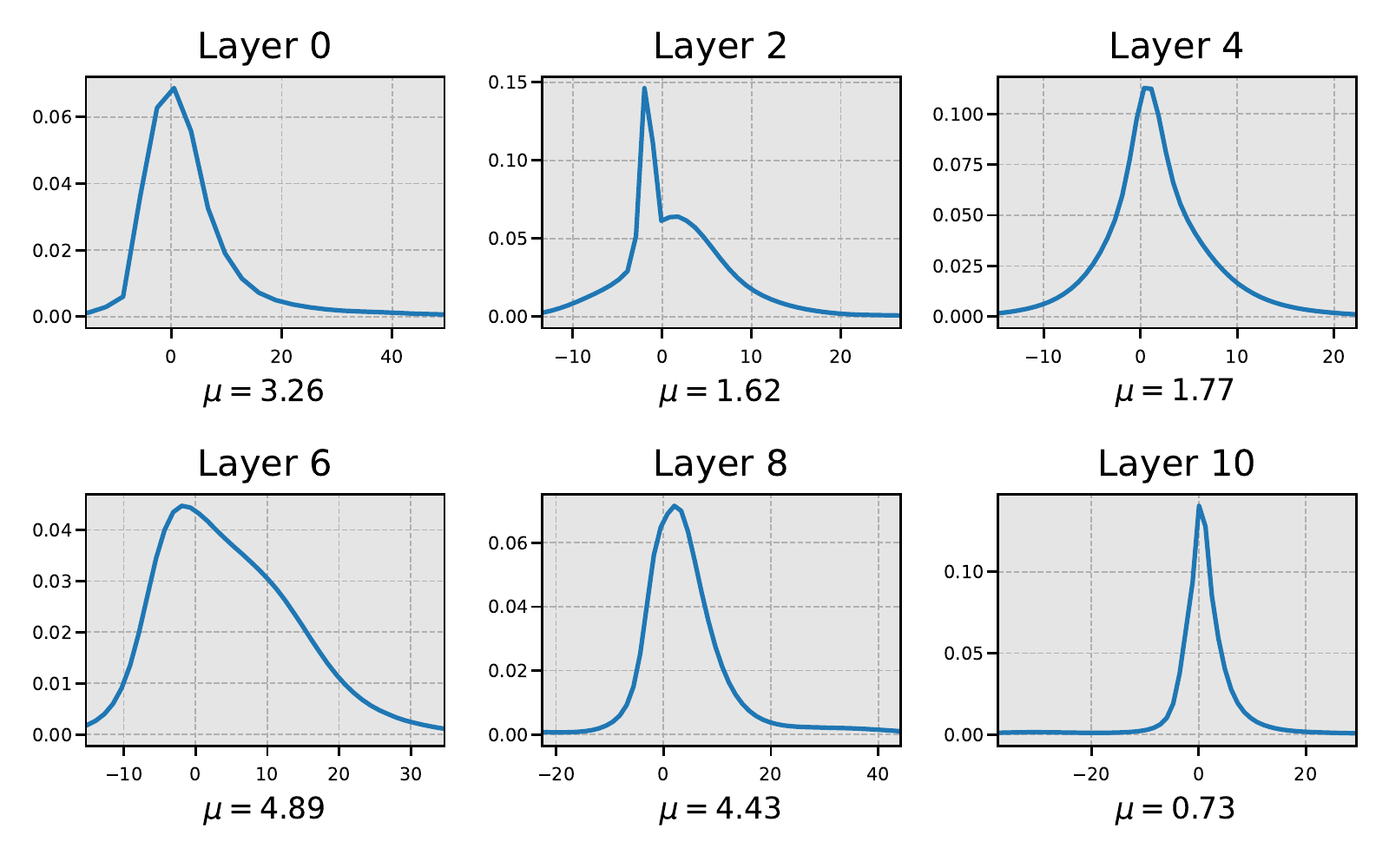}}
	\caption{Probability density function of the distribution of values of the $Q K^\top$ matrix at different layers of a trained transformer using \name, showing the middle 99\% of values. $\mu$ is the mean of each distribution. We find the distributions to be approximately centered around zero.
	}
	\label{fig:qk_distr}
\end{figure}

We center the Taylor expansion of the exponential around zero, i.e. we use the Maclaurin series, due to mathematical considerations and practical implications.
The Taylor series of the exponential function around the point $a \in \mathbb R$ is
$$
\exp(x) \approx \exp(a) \left( 1 + (x - a) + \frac 1 2 (x - a)^2 \right).
$$
One can see that the difference when choosing different points of expansion $a$ is just a shift (addition) of the input and scaling (multiplication) of the output.
These operations are naturally adjusted by the network during training.

We further justify this choice by considering the normalization of queries and keys, which are scaled by the learnable attention temperature.
As these values are constrained to lie on a sphere centered around zero, the entries of the resulting $QK^\top$ matrix are also centered around zero.
Moreover, empirical analysis of a  trained TaylorShift transformer on ImageNet confirms that the activations of $QK^\top$ are indeed approximately centered around zero, as illustrated in \Cref{fig:qk_distr}. This empirical evidence supports the effectiveness of centering the Taylor expansion at zero.

Additionally, we acknowledge the practical constraints that would be associated with dynamically adjusting the centering point of the Taylor approximation, as the point of efficient TaylorShift is not explicitly computing the $QK^\top$ matrix.

\subsection{Empirical and Theoretical Efficiency Transition Points}
\label{sec:emp_theo_diff_details}

The difference between the theoretical and empirical transition points $N_0$ and $\hat N_0$ in Figure 2 of the main paper
hints at possible gains in speed for efficient \name especially, since it shows that currently, our implementation runs at fewer FLOPS per second than direct \name.
This indicates that efficient \name is memory bound by saving and loading large intermediate results like $A_\text{mod}$.
This might be complicated by the increased internal dimension from $d$ to $d^2$.
We hypothesize, however, that it should also be possible to apply a strategy similar to Flash \cite{Dao2022} to only compute parts of $A_\text{mod}$ at a time, respecting the GPUs memory hierarchy.

\subsection{Varying Sequence Length $N$}
\begin{figure}[t]
	\centering
	\includegraphics{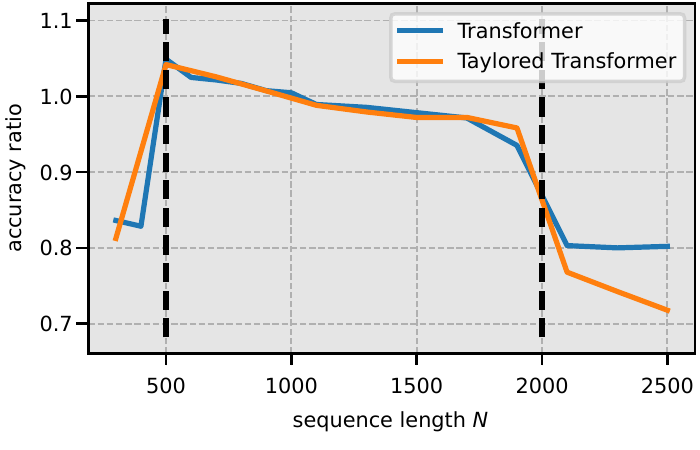}
	\caption{Ratio between the accuracy obtained on the test set and at a specific sequence lengths of the ListOps task.
		The training and test sets contain sequences of length 500 to 2000, marked by the black dashed lines.
	}
	\label{fig:evaluate_sequence_length}
\end{figure}
We explore the performance of our model on sequences of varying length in \Cref{fig:evaluate_sequence_length}.
For both the baseline and \name, accuracy gradually declines within the training distribution, spanning from 500 to 2000 tokens.
We attribute this trend to the increasing complexity of solving mathematical operations as the number of operations grows.
Outside the training distribution, accuracy drops rapidly to approximately $80\%$ of the test accuracy, with the accuracy of \name decreasing slightly more in this out-of-distribution setting.

\subsection{Efficiency Comparison}
\label{sec:appendix_eff_comparison}

\begin{table}[ht!]
	\centering
	\caption{Training speed and memory requirements for the transformers from Table 3 of the main paper.
	Hyperparameters can be found in \Cref{tab:training_hyperparameters}.}
	\resizebox{\textwidth}{!}{
		\begin{tabular}{l*4{S[table-format=3.2]}*4{S[table-format=3.2]}}
			\toprule
			\multirow{2}{*}{Model} & \multicolumn{4}{c}{training speed [h*GPUs]} & \multicolumn{4}{c}{training memory [GB]} \\
			& {CIFAR} & {IMDB} & {ListOps} & {ImageNet (S)} & {CIFAR} & {IMDB} & {ListOps} & {ImageNet (S)} \\
			\cmidrule(r){1-1} \cmidrule(lr){2-5} \cmidrule(l){6-9}
			Linformer & 2.01 & 7.10 & {-} & 91.35 & 5.50 & 9.83 & {-} & 152.46 \\
			RFA 		& 2.51 	& 7.88 & {-} & {-} & 7.64 & 9.09 & {-} & {-} \\
			Performer & 2.91$^\star$ & 9.51$^\star$ & 46.17$^\star$ & 141.06$^\star$ & 9.15$^\star$ & 11.20$^\star$ & 66.14$^\star$ & 198.10$^\star$ \\
			Reformer & 5.83 & 34.74 & 103.11 & 622.82$^\star$ & 28.83 & 44.34 & 173.67 & 378.96$^\star$ \\
			Nystromformer & 2.07 & 8.01 & 40.66 & 196.02$^\star$ & 5.13 & 8.55 & 30.92 & 266.56$^\star$ \\
			EVA 		& 2.64 & 9.20 & 52.57 & 147.03 & 4.82 & 8.59 & 58.87 & 124.34 \\
			\cmidrule(r){1-1}
			Transformer & 2.15 & 18.05 & 48.50 & 87.45 & 11.95 & 58.21 & 273.19 & 107.01 \\
			\cmidrule(r){1-1}
			direct TaylorShift & 2.85 & 24.06 & 161.99 & 96.46 & 16.19 & 78.94 & 545.22 & 132.32 \\
			efficient TaylorShift & 4.58 & 25.48 & 208.67 & {-} & 26.53 & 39.50 & 401.87 & {-} \\
			\bottomrule
	\end{tabular}}
\end{table}

\begin{figure}[ht!]
	\centering
	\includegraphics[width=\textwidth]{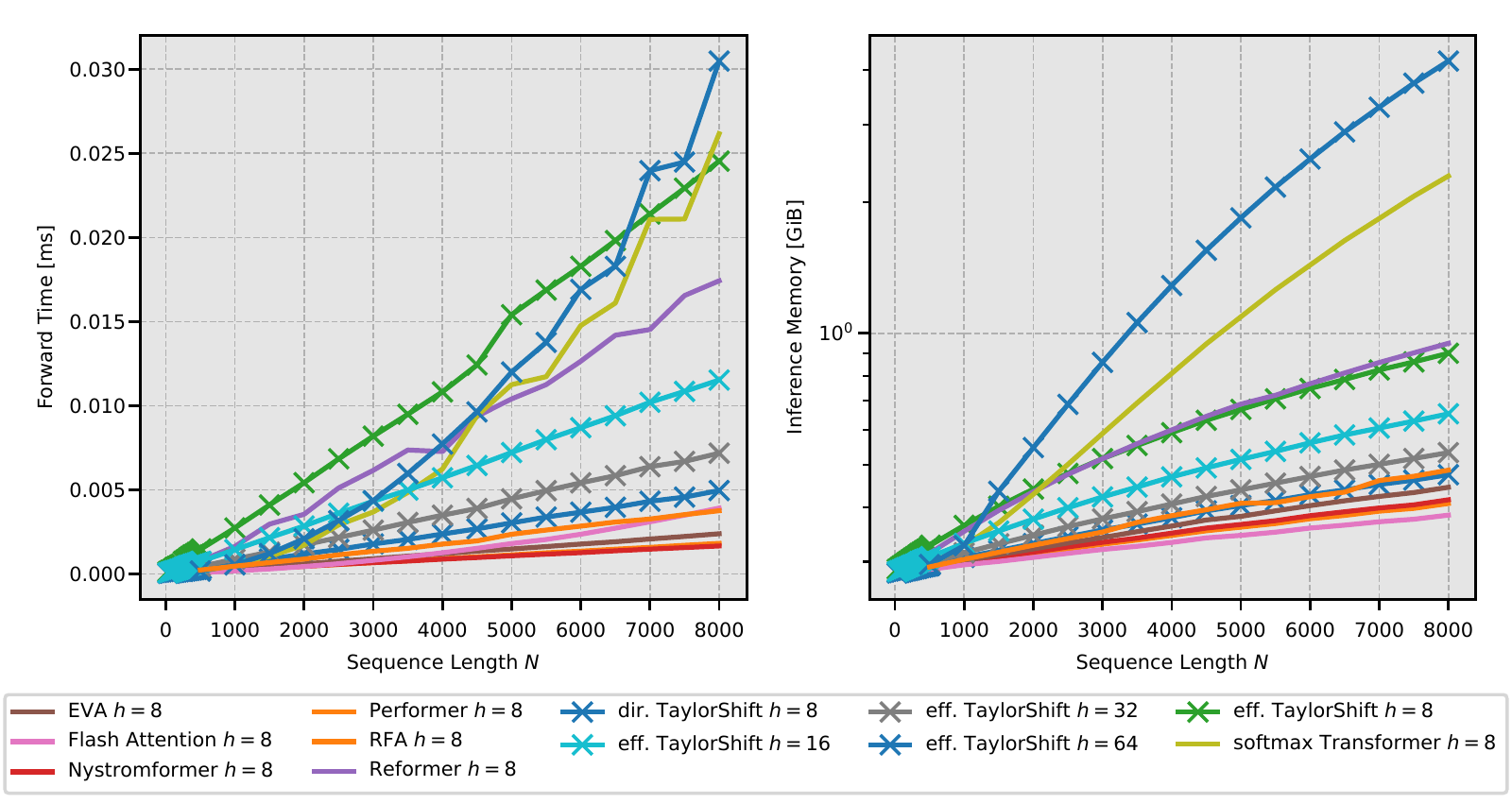}
	\caption{Inference time and memory for full transformer models using different attention mechanisms. Experiments are run using the hyperparameters from the ListOps dataset.
	}
	\label{fig:real_world_mem_and_tp_extended}
\end{figure}
We compare the full inference model speed and memory requirements in \Cref{fig:real_world_mem_and_tp_extended}, which extends Figure 3 of the main paper,
including more models.
While at the default number of attention heads $h = 8$ with per-head embedding dimension $d = 64$ \name is not competitive with other efficient attention mechanisms, we have shown in Table 6 of the main paper
that one can increase the number of heads, reducing the per-head dimension down to $d = 16$ without loss of accuracy (in fact, we increase accuracy by increasing the number of heads) or even $d = 8$ with only a minor drop in accuracy.
Utilizing this fact and increasing the number of heads to $h = 32$ or even $h = 64$ makes \name very competitive with other efficient attention mechanisms and demonstrates it's superior scaling.

\subsection{Token Embedding}
\begin{table}[t]
	\centering
	\caption{Accuracy on different datasets when changing the token embedding from a linear layer to a 3-layer CNN.}
	\label{tab:ablations_embedding}
	\begin{tabular}{lccc}
		\toprule
		Dataset & lin. embed. & conv. embed. & $\Delta$ \\
		\cmidrule(r){1-1} \cmidrule(l){2-4}
		CIFAR (Pixel) & 47.1 & 51.1 & 4.0 \\
		IMDB (Byte) & 66.0 & 86.3 & 20.3 \\
		ListOps & 45.6 & 64.8 & 19.2 \\
		ImageNet (Ti) & 75.0 & 77.1 & 2.1 \\
		ImageNet (S) & 79.3 & 78.5 & -0.8 \\
		\bottomrule
	\end{tabular}
\end{table}
To test an orthogonal angle influencing efficiency, we take a look at the initial token embedding fed into a \name-equipped Transformer encoder.
Table \ref{tab:ablations_embedding} contrasts accuracy when transitioning from linear token embedding to a 3-layer CNN\footnote{1D for CIFAR, IMDB, and ListOps and 2D for ImageNet}.
Notably, incorporating the CNN-embedding yields large performance improvements in the sequence-based tasks, indicating a complementing effect of convolutions and \name.
We did not employ the CNN-embedding in other experiments to preserve experimental comparability.
However, including it is an easy and efficient way of increasing model performance with linear complexity, without having to change the whole backbone architecture.

\end{document}